\documentclass{article} %
\usepackage{iclr2025,times}

\usepackage[utf8]{inputenc} %
\usepackage[T1]{fontenc}    %
\usepackage{hyperref}       %
\usepackage{url}            %
\usepackage{booktabs}       %
\usepackage{amsfonts}       %
\usepackage{nicefrac}       %
\usepackage{microtype}      %
\usepackage{xcolor}         %

\usepackage{amsmath}
\usepackage{bm}
\usepackage{amsthm}

\usepackage{subcaption}
\usepackage{wrapfig}
\usepackage{graphicx}
\usepackage[inline]{enumitem}
\usepackage{soul}
\usepackage{comment}

\usepackage{algorithm}
\usepackage{algpseudocode}

\usepackage{xcolor}

\title{Attention as a Hypernetwork}

\author{%
  Simon Schug \\
  ETH Z\"{u}rich\\
  \texttt{sschug@ethz.ch} \\
  \And
  Seijin Kobayashi \\
  ETH Z\"{u}rich\\
  \texttt{seijink@ethz.ch} \\
  \AND
  Yassir Akram \\
  ETH Z\"{u}rich\\
  \texttt{yakram@ethz.ch} \\
  \And
  João Sacramento\thanks{Shared senior authors} \\
  {\footnotesize Google, Paradigms of Intelligence Team} \\
  \texttt{joaosacramento@google.com} \\
  \And
  Razvan Pascanu$^\dag$ \\
  Google DeepMind \\
  \texttt{razp@google.com} \\
}

\iclrfinalcopy %
\begin{document}
\maketitle
\setcounter{footnote}{0}
\begin{abstract}
Transformers can under some circumstances generalize to novel problem instances whose constituent parts might have been encountered during training, but whose compositions have not.
What mechanisms underlie this ability for compositional generalization?
By reformulating multi-head attention as a hypernetwork, we reveal that a composable, low-dimensional latent code specifies key-query specific operations.
We find empirically that this latent code is predictive of the subtasks the network performs on unseen task compositions, revealing that latent codes acquired during training are reused to solve unseen problem instances.
To further examine the hypothesis that the intrinsic hypernetwork of multi-head attention supports compositional generalization, we ablate whether making the hypernetwork-generated linear value network nonlinear strengthens compositionality.
We find that this modification improves compositional generalization on abstract reasoning tasks.
In particular, we introduce a symbolic version of the Raven's Progressive Matrices human intelligence test, which gives us precise control over the problem compositions encountered during training and evaluation.
We demonstrate on this task how scaling model size and data enables compositional generalization in transformers and gives rise to a functionally structured latent space.
\footnote{Code available at \url{https://github.com/smonsays/hypernetwork-attention}}
\end{abstract}

\section{Introduction}
\label{sec:introduction}
Abstract reasoning is a cornerstone of human intelligence.
Many intelligence tests in psychology measure intelligence using abstract reasoning tasks \citep[e.g.][]{wechsler_measurement_1958, binet_development_1961, raven_advanced_1962}.
Arising from the idea that combinatorial structures are central to human cognitive abilities, the ability of connectionist models to reason has been debated for decades \citep[e.g.][]{rumelhart_learning_1986, fodor_connectionism_1988, smolensky_connectionism_1991}.
However, with the success of neural networks at scale, the capacity for reasoning has seemingly come into reach \citep{huang_towards_2023}.
Arguably key to this success is the ability for \emph{in-context learning} paired with the capacity to flexibly recombine learned skills and knowledge to unseen settings.
While these capabilities emerge at least partially from training large models on huge datasets \citep{brown_language_2020}, it is not clearly understood how the resulting networks implement them - and why they still often fail.

A number of recent studies that aim to illuminate in-context learning abilities have found that transformers can learn to perform gradient-based optimization within sequence \citep{dai_why_2023, akyurek_what_2023, von_oswald_uncovering_2023}.
They build on the insight that linear attention can be understood as a fast-weight programmer \citep{schmidhuber_learning_1992},
that learns to construct an implicit model in-context, with weights given by a sum of input-dependent outer products \citep{schlag_linear_2021}. 
Complementing this perspective, \citep{hendel_-context_2023, todd_function_2024} have found that in-context learning creates task vectors in the hidden activations of deeper layers of the network, which summarize in-context information and modulate the forward computation.
What is striking is that learning in-context can under some circumstances lead to \textit{compositional generalization}, that is generalization to unseen combinations of previously observed constituents \citep{an_how_2023, lake_human-like_2023, hosseini_compositional_2022}, a capacity that neural networks notoriously struggle with \citep{srivastava_beyond_2023, press_measuring_2023, dziri_faith_2023}.
The mechanism behind this ability however is not well understood.
Here we aim to shed light on this question.

\begin{figure}[t]
    \centering
    \includegraphics[width=\textwidth]{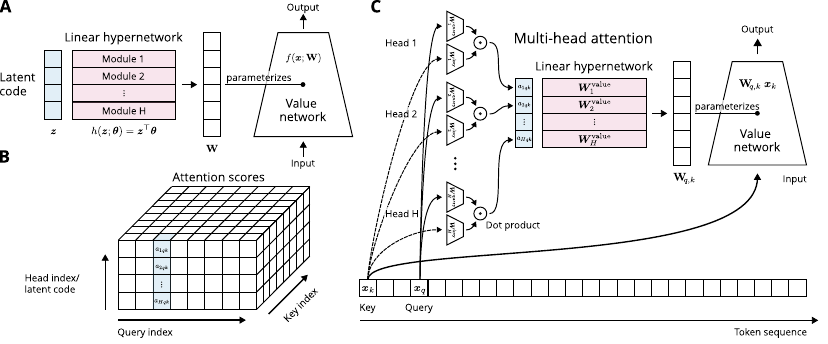}
    \caption{\textbf{Hypernetwork attention}. \textbf{A} A linear hypernetwork maps a latent code to a set of parameters that configure a value network to process the input. \textbf{B} The attention scores along the head index form the latent code of the hypernetwork. \textbf{C} Multi-head attention can be equivalently expressed as a linear hypernetwork that configures key-query specific computations of a linear value network.
    }
    \label{fig:hypernetwork-attention}
    \vspace{-1em}
\end{figure}

Our key finding is that through the use of multiple heads, the attention mechanism of transformers is mathematically equivalent to a hypernetwork, a neural network that reconfigures the computation of another neural network as shown in Figure~\ref{fig:hypernetwork-attention}.
This results in a novel interpretation of the attention scores along the head index as a compact latent code that specifies the operations the attention layer applies to each input.
Notably, we hypothesize that because multi-head attention shares the same hypernetwork per layer, it is encouraged to reuse and recombine previously acquired operations.

In order to investigate whether this constitutes a faithful characterization of what transformers learn in practice, we study in-context learning tasks that require recomposing knowledge obtained during training.
As one such task, we develop a challenging abstract reasoning task based on the Raven's Progressive Matrices human intelligence test \citep{raven_advanced_1962} and show how multi-head attention develops a structured latent code that captures information about the operations implemented.

Our \textbf{main contributions} can be summarized as follows:
\begin{itemize}[leftmargin=*]
  \item We reformulate standard multi-head attention from a hypernetwork perspective, revealing that network computations are specified by a low dimensional latent code.
  \item We show that scaling up model size and data enables transformers to compositionally generalize on abstract reasoning tasks and leads to a structured latent space that is predictive of network function.
  \item To test the hypothesis that the hypernetwork mechanism supports compositionality, we introduce a simple modification to multi-head linear attention that makes the value network nonlinear without introducing additional parameters, finding that it improves compositional generalization.
  \item We introduce \textsc{sraven}, a challenging symbolic, abstract reasoning task whose difficulty can be controlled parametrically and which offers precise control over the problem compositions encountered during training in order to test for compositional generalization.
\end{itemize}

\section{Attention as a hypernetwork}
\label{sec:attention-as-hypernetwork}

In this section, we will first briefly recapitulate hypernetworks \citep{ha_hypernetworks_2017} and standard multi-head dot-product attention \citep{vaswani_attention_2017} before showing how attention can equivalently be expressed as a hypernetwork.
We will then introduce Hypernetwork Linear Attention (HYLA) as a simple modification of linear attention that renders the hypernetwork mechanism inside attention more expressive.

For notation, we will use bold lower-case letters for vectors (e.g. $\bm{z}$), bold upper-case letters for matrices (e.g. $\bm{\mathrm{A}}$) and bold, italic upper-case letters for learnable parameter matrices (e.g. $\bm{W}_V$).

\subsection{Multi-head attention}

A \textit{hypernetwork} is a neural network $h(\bm{z}; \bm{\theta})$ parameterized by $\bm{\theta}$ that takes as input a \textit{latent code} $\bm{z}$ and outputs parameters $\bm{\mathrm{W}}$ that parameterize a \textit{value network} $f(\bm{x}; \bm{\mathrm{W}})$, which then processes the inputs (see Figure~\ref{fig:hypernetwork-attention}A).
The latent code $\bm{z}$ is typically low-dimensional and can be interpreted as a specification of the computation performed by the value network.
As we will see in the following, multi-head attention can be viewed as a linear hypernetwork producing the weights for a linear value network.
Note that this is different from the observation that linear attention can be understood as a fast weight programmer \citep{schlag_linear_2021}, where the fast weights of a value network are constructed as a sum of outer products over the \textit{key indices} instead of the \textit{head indices}.

Self-attention\footnote{We focus on self-attention, but the same derivation holds for cross-attention, where the keys and values are obtained as the output of an encoder and the queries come from a decoder.} maps a sequence of inputs $\bm{\mathrm{X}} \in \mathbb{R}^{D \times T}$ to a sequence of outputs $\bm{\mathrm{Y}} \in \mathbb{R}^{D \times T}$.
For each attention head $h \in \{1, \ldots, H\}$, the inputs are projected into keys $\bm{\mathrm{K}}_h = \bm{W}^{\text{key}}_h \bm{\mathrm{X}}$ and queries $\bm{\mathrm{Q}}_h =  \bm{W}^{\text{query}}_h \bm{\mathrm{X}}$ using head-specific projection matrices $\bm{W}^{\text{key}}_h, \bm{W}^{\text{query}}_h, \bm{W}^{\text{value}}_h \in \mathbb{R}^{D^{\text{head}} \times D}$ with $D^{\text{head}} = \frac{D}{H}$.
The attention matrix can then be obtained as
\begin{align}
\bm{\mathcal{A}} = \sigma \bigl( \; \left[\bm{\mathrm{\tilde{A}}}_1 \; \bm{\mathrm{\tilde{A}}}_2 \; \ldots \; \bm{\mathrm{\tilde{A}}}_H \right] \; \bigr ) \text{ with } \bm{\mathrm{\tilde{A}}}_h =  \frac{\bm{\mathrm{Q}}_h^\top \bm{\mathrm{K}}_h}{\sqrt{D^{\text{head}}}}  
\end{align}
where $\bm{\mathcal{A}}=(a_{h,q,k})_{h,q,k} \in \mathbb{R}^{H \times T \times T}$ stacks the head-specific unnormalized attention matrices~$(\bm{\mathrm{\tilde{A}}}_h)_h$ into a three-dimensional array and applies a normalization operation $\sigma(\cdot)$.
For linear attention, $\sigma(\cdot)=\mathrm{Id}(\cdot)$ is simply the identity whereas for softmax attention $\sigma(\cdot)=\mathrm{Softmax}(\cdot)$ applies the softmax operation for each head and query index independently across the key indices.
Considering a single query index $q \in \{1, \ldots, T\}$ and $\mathbf{x}_k \in \mathbb{R}^D$ the $k$-th column of $\bm{\mathrm{X}}$, multi-head attention (MHA) can be equivalently expressed as a hypernetwork:
\begin{align}
\text{MHA}_{q}(\bm{\mathrm{X}})
:=& \bm{W}^{\text{out}} \bigoplus_{h=1}^H \sum_{k=1}^T a_{h,q,k} \; \bm{W}^{\text{value}}_h \mathbf{x}_k \\
=& \sum_{h=1}^H \bm{W}^{\text{out}}_h \sum_{k=1}^T a_{h,q,k} \; \bm{W}^{\text{value}}_h \mathbf{x}_k \\
=& \sum_{k=1}^T \left(\sum_{h=1}^H \underbrace{a_{h,q,k}}_{\text{latent code}} \; \underbrace{\bm{W}^{\text{out}}_h \bm{W}^{\text{value}}_h}_{\text{hypernetwork}} \right) \mathbf{x}_k \label{eq:hya-main} \\
=& \sum_{k=1}^T \underbrace{\bm{\mathrm{W}}_{q,k}}_{\text{value network}} \mathbf{x}_k \label{eq:hya-last}
\end{align}
where $\oplus$ denotes concatenation and the $\bm{W}^{\text{out}}_h \in \mathbb{R}^{D\times D^{\text{head}}}$ are defined as the slices of $\bm{W}^{\text{out}} \in \mathbb{R}^{D\times D}$ such that $\bm{W}^{\text{out}} = \bigoplus_{h=1}^H \bm{W}^{\text{out}}_h$.
See Figure~\ref{fig:hypernetwork-attention}C for a diagram of the computation.

As a result, the key-query specific attention scores over the multiple heads form a latent code, where the number of heads $H$ corresponds to the \textit{latent dimension} of a hypernetwork.
Accordingly, the dot products between any key-query pair over all heads can be interpreted as an amortized inference process to configure the computation implemented in the linear value network.

The key significance of our finding is that it suggests that the use of multiple heads in attention allows it to implement specialized, reusable operations through the latent code of a hypernetwork.
It has been theoretically and empirically shown that hypernetworks support compositional generalization \citep{schug_discovering_2024}, providing a possible explanation for similar capabilities observed in transformers \citep{lake_human-like_2023, an_how_2023, hosseini_compositional_2022}.
Note that multi-head attention reuses the same hypernetwork for every key-query index.
Specifically, the hypernetwork linearly combines the \emph{same} matrices $(\bm{W}^{\text{out}}_h, \bm{W}^{\text{value}}_h)_h$ according to the weights given by $a_{h,q,k}$.
This incentivizes reuse of latent codes and their associated operations implemented through the value network.

\subsection{Hypernetwork Linear Attention (HYLA)}

If the perspective of multi-head attention as a hypernetwork indeed reflects how it operates in practice, it should be possible to modify multi-head attention in a way that further reinforces the hypernetwork mechanism.
For instance, both the hypernetwork and the value network in multi-head attention are simple linear networks and could be replaced by nonlinear networks.
Furthermore, the normalization operation $\sigma(\cdot)$ could be used to encode inductive biases such as competition or sparsity in the latent code by imposing structural constraints.

Here, we focus on a simple modification to linear attention, where we make the value network nonlinear and normalize the latent code along the head index (instead of along the key index as in standard softmax attention).
Specifically, we define Hypernetwork Linear Attention (HYLA) to use a single hidden layer value network by adding a nonlinearity.
This should increase the expressivity of the subfunctions that are learned and recomposed by the layer.
Noting that the output and value projections $\bm{W}^{\text{out}}_h,  \bm{W}^{\text{value}}_h$ already form a deep linear network in standard multi-head attention (c.f. Eq.~\ref{eq:hya-main}) this can be achieved without adding additional parameters as follows:
\begin{align}
\text{HYLA}_{q}(\bm{\mathrm{X}})
=& \sum_{k=1}^T \left(\sum_{h=1}^H a_{h,q,k} \; \bm{W}^{\text{out}}_h  \right) \phi \left ( \sum_{h=1}^H a_{h,q,k} \; \bm{W}^{\text{value}}_h \mathbf{x}_k \right ) \\
=& \sum_{k=1}^T \bm{\mathrm{W}}'_{q,k} \phi (\bm{\mathrm{W}}_{q,k} \mathbf{x}_k),
\end{align}
where $\phi(x)$ is an element-wise nonlinearity; here we set it to be a ReLU, $\phi(x) = \max(0, x)$.
Additionally, we use $\sigma(\cdot) = \text{RMSHead}(\cdot)$, to normalize the attention scores for each key and query index independently across the head indices using $\text{RMSNorm}(\bm{x})=\frac{\bm{x}}{\sqrt{\frac{1}{n}\sum_{i=1}^{n}x_{i}^{2}}}$ (without adding a learnable scalar as in \citet{zhang_root_2019}).
This ensures that the parameters of the value network generated by the hypernetwork maintain the variance preserving properties of neural network initializations that have been found to be important for the stability of gradient-based training \citep{glorot_understanding_2010}.
The resulting HYLA-layer is a simple drop-in replacement to standard attention.
Notably, since the normalization operation $\sigma(\cdot)$ is local to the query and key index, it requires no communication across keys as opposed to softmax attention.
By making the value network more expressive, we allow the network to implement more complex operations for a given latent code which we hypothesize strengthens the ability for compositional generalization.

\section{Compositional generalization on fuzzy logic functions}
\label{sec:fuzzy-logic}
\begin{figure}[t]
    \centering
    \includegraphics[width=\textwidth]{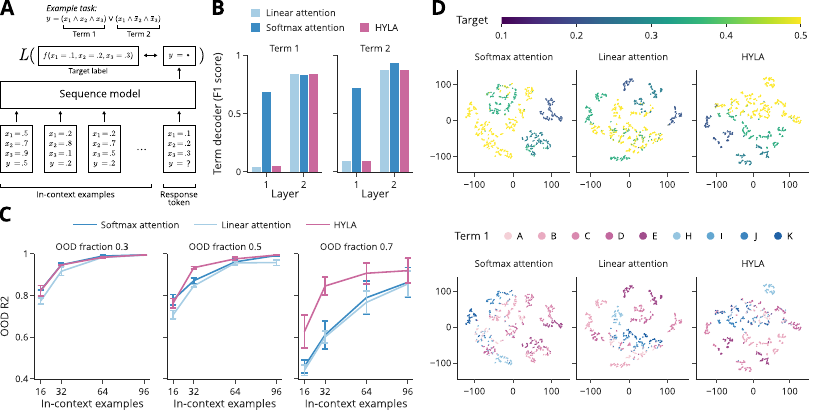}
    \caption{\textbf{Compositional generalization on fuzzy logic functions}. \textbf{A} We split fuzzy logic functions according to their constituent terms into train and out-of-distribution (OOD) sets to measure compositional generalization in a sequence model that learns these functions in-context. \textbf{B} The latent code of the response token is predictive of the constituent terms underlying each task. Shown is the F1 score on unseen tasks of logistic regression classifiers for each layer and term, trained to predict the terms underlying each task based on the attention scores across the head index for the response token attending to itself. \textbf{C} Task performance on unseen tasks reported as OOD $R^2$ for varying number of in-context examples and fraction of tasks held-out during training. \textbf{D} tSNE visualization of the latent codes (attention scores across the head index for the response token attending to itself) colored according to the target label (top) and colored according to the first term of the fuzzy logic function of each task (bottom).}
    \label{fig:flogic}
    \vspace{-1em}
\end{figure}

We start by developing a fuzzy logic task with compositional structure, inspired by a similar task previously considered by \citet{rahaman_dynamic_2021}.
Learning such tasks in an in-context learning setting will allow us to study the ability of multi-head attention-based models to compositionally generalize and analyze the structure of their latent code when solving each task.

Given a set of $L$ scalars $\{x_1, ..., x_L \}$ with $x_i \in [0, 1]$ we define fuzzy logic operators on the elements of this set as the Zadeh operators \citep{zadeh_fuzzy_1965}:
\begin{align}
  x_i \land x_j := \min(x_i, x_j), \quad x_i \lor x_j := \max(x_i, x_j), \quad \bar{x}_i := 1 - x_i 
\end{align}
These operators have the property that for boolean inputs $x_i \in \{0, 1\}$ the outputs coincide with the outputs of the corresponding Boolean operation.
Each task is then defined as a function in disjunctive normal form consisting of $K$ terms such as for example for $L=4$ and $K=2$
\begin{align}
  f(x_1, x_2, x_3, x_4) = \underbrace{(x_1 \land x_2\land x_3 \land x_4)}_{\text{Term 1}} \lor \underbrace{(\bar{x}_1 \land x_2\land \bar{x}_3 \land x_4)}_{\text{Term 2}}
\end{align}
where each term is a conjunction of all variables with negations applied to some of them.
In order to test for compositional generalization, we index each of the $2^L$ possible terms, generate all possible combinations of $K$ terms and hold out a fraction of the combinations for out-of-distribution (OOD) testing, ensuring that their constituent terms have been encountered in some tasks during training.
For each task, we present $N$ examples in-context (see Figure~\ref{fig:flogic}A) where each token has dimension $L+1$, consisting of the concatenated inputs sampled uniformly from $[0,1]$ and the corresponding target of the function for the current task.
For the final token, we mask out the target and use the output logits of the final token to predict the target, measuring the loss using the mean-squared error.

\subsection{Compositional generalization}

We first investigate how well multi-head attention based transformer models are able to learn our fuzzy logic tasks in-context and compositionally generalize to held-out tasks.
We vary the number of examples presented in context, as well as the fraction of combinations of fuzzy logic terms held-out from the training set.
Figure~\ref{fig:flogic}C compares standard multi-head softmax attention and linear attention, as well as our HYLA model in terms of their coefficient of determination ($R^2$) on held-out tasks.
We find that all models are able to solve the fuzzy logic task given sufficient examples are presented in context.
As the fraction of held-out tasks increases, compositional generalization performance starts decreasing for all models.
Interestingly, HYLA is less affected by this decline than softmax and linear attention.
We hypothesize that the more expressive value network allows to more efficiently represent the nonlinear terms being composed and thereby strengthens the inductive bias towards compositionality.
Consistent with this interpretation, in-distribution the models show comparably small differences in performance (see Figure~\ref{appfig:flogic-seq-len-frac-test}B+D).
In Appendix~\ref{appsec:ablation} we conduct additional model ablations that separate the contributions of the RMSHead normalization and the nonlinearity added to the value network, showing that both components in combination are required to obtain the observed performance improvements.

Our compositional split only contains novel combinations of \textit{known} terms.
A priori, it might be possible that the system learns to compose the basic fuzzy logic operations in a way that allows it to generalize to any fuzzy logic function, including those that contain novel combinations of \textit{unknown} terms.
However, testing on functions obtained as novel combinations of $K$ unknown terms, we find that none of the models considered here is able to solve such a task (see Figure~\ref{appfig:flogic-seq-len-frac-test}C).
\vspace{-1ex}
\subsection{Latent code structure}
Next, we analyze the structure of the latent code for the different models solving the task.
As suggested by Eq.~\ref{eq:hya-main}, the attention scores for a given key-query pair across the head dimension configure the computation performed in the value network, and accordingly we would expect it to reflect the fuzzy logic function that needs to be implemented to correctly predict the output of the response token.
To test this hypothesis, we collect the attention scores for the response token attending to itself, obtaining a vector of dimension $H$ for each layer and task.
For this purpose, we use the tasks held-out from training, varying the response token while keeping inputs shown in-context fixed.
In order to visualize the $H$-dimensional points we thereby obtain for each layer, we reduce the dimensionality using tSNE \citep{maaten_visualizing_2008}.
The results of this analysis are shown in Figure~\ref{fig:flogic}D (for an extended figure showing all layers, see Figure~\ref{appfig:flogic-latents}).
In line with our hypothesis, the latent codes for each model form clusters.
The structure of these clusters is only partially explained by the specific output value of each function.
Strikingly, coloring data points according to the fuzzy logic terms that make up each function reveals that clusters in the latent codes correspond to the terms.
Indeed, it is possible to decode the terms underlying each function from the latent code using a logistic regression classifier which is trained to predict the function terms given the latents and is evaluated on held-out tasks as shown in Figure~\ref{fig:flogic}B.

\section{\textsc{sraven}: Symbolic raven}
\label{sec:sraven}
\begin{figure}[t]
    \centering
    \includegraphics[width=\textwidth]{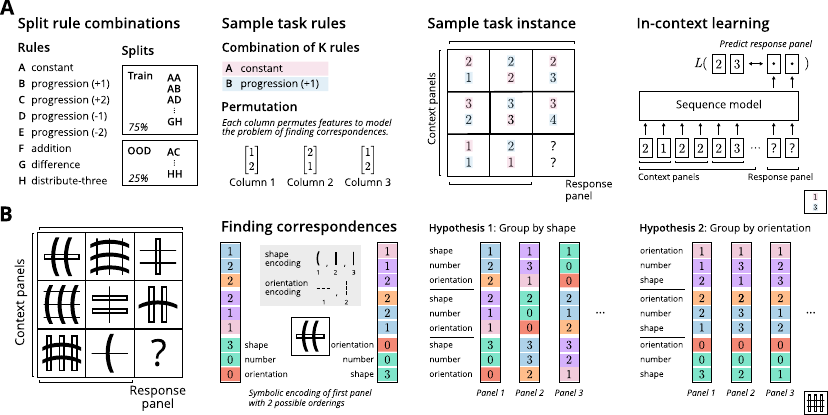}
    \caption{\textbf{\textsc{sraven}}. \textbf{A} Illustration of \textsc{sraven} task generation and the construction of the compositional generalization split. \textbf{B} Example problem instance illustrating a key challenge of the original Raven's Progressive Matrices of \textit{finding correspondences} (adapted from \citet{carpenter_what_1990}). When attempting to solve this instance, different hypotheses over which figural elements are governed by a consistent rule across rows are possible. This is akin to different orderings of the symbolic features.}
    \label{fig:sraven-task}
    \vspace{-1em}
\end{figure}
Next, we introduce an abstract reasoning task based on Raven's Progressive Matrices \citep{raven_advanced_1962}, we refer to as \textsc{sraven}.
We develop this task to provide a challenging benchmark that probes symbolic reasoning capabilities while giving us fine-grained control over the constituent parts of each task in order to assess compositional abilities and analyze latent code structure. 

\subsection{Abstract reasoning based on symbolic Raven's Progressive Matrices}
Raven's Progressive Matrices are a classic human intelligence test that probes the ability to induce abstract relations \citep{raven_advanced_1962}.
It requires test subjects to generate and verify a potentially large number of hypotheses over rules that can parsimoniously explain a given problem instance \citep{carpenter_what_1990}.
While there have been previous machine learning datasets inspired by Raven's Progressive Matrices~\citep{wang_automatic_2015, zhang_raven_2019, barrett_measuring_2018, hu_stratified_2021}, we seek to study a setting that is both symbolic, compositional and models a key difficulty of Raven tasks referred to as \textit{finding correspondences} by \citet{carpenter_what_1990} which requires searching through a large number of possible hypotheses.
We discuss existing Raven-based benchmarks in Section~\ref{sec:related-work}.

\subsection{Task description}
Figure~\ref{fig:sraven-task}A illustrates the generative process of \textsc{sraven}.
For each task, a matrix of eight context panels is presented, and the goal is to predict the contents of the final response panel.
In the original Raven's Progressive Matrices Test, each panel typically displays a combination of geometrical shapes, as for instance shown in Figure~\ref{fig:sraven-task}B.
For our symbolic version, each panel is defined as a tuple of $K$ integers that symbolically encode features (in our experiments we set $K=4$ unless stated otherwise).
The $K$ features within each row evolve according to $K$ \textit{rules}.
Every rule is a function that takes two
integers as input and outputs an integer.
Figure~\ref{fig:sraven-task}A lists all rules, and Appendix~\ref{appsec:sraven-rules} describes each rule in detail.
In order to maintain a finite set of possible symbols, we limit all tasks to contain only integers $\{0, 1, 2, \ldots, F-1\}$ (in our experiments, we set $F=8$).
Arithmetic rules such as progression, addition or difference are therefore defined using modular arithmetic modulo $F$.

To create a task, we first sample $K$ out of the $R=8$ possible rules and then use each of the $K$ rules, to generate three sequences of length three given random inputs to each rule.
As a result, we obtain nine panels, where each row can be explained by the same set of $K$ underlying rules.
We use the first eight panels as context panels and present them sequentially to a sequence model, which needs to predict each feature of the response panel independently.
A prediction is considered correct if and only if all subpredictions were correct.
In the original Raven task, a number of possible answer panels are presented alongside the context panels and test subjects are asked to pick the one which contains the correct answer \citep{raven_advanced_1962}.
Since in our case, the correct answer will consist of a combination of a finite set of known symbols, we can ask the agent solving the task to directly predict the missing symbols.
Conveniently, this helps avoid potential shortcuts for solving a task that might arise from a biased generative process to create multiple-choice answers as a popular prior Raven-based benchmark has been found to be affected by \citep{zhang_raven_2019, hu_stratified_2021}.

One of the key features of \textsc{sraven} is its compositional structure.
Each task consists of a combination of a finite set of rule combinations.
To systematically test whether a system trained on a subset of rule combinations can generalize to unseen combinations, we split all rule combinations into a train set and a test set.
Unless noted otherwise, we hold out $25\%$ of all possible rule combinations for evaluation in our experiments.
Note that permutations of rule combinations (e.g. AB and BA) are considered to be the same and will only appear in one of the two sets.
This ensures that at test time, the agent is confronted with instances of tasks whose underlying combination of rules it has never encountered during training.

\subsection{Finding correspondences}
\label{sec:raven-finding-correspondences}
\begin{figure}[t]
    \centering
    \includegraphics[width=\textwidth]{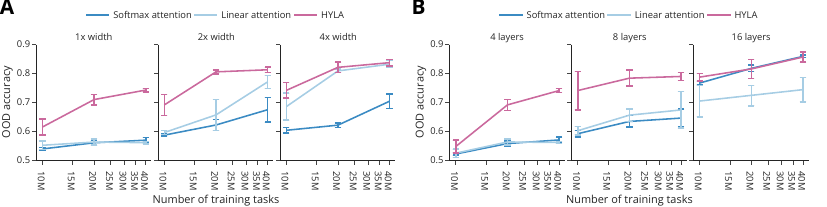}
    \caption{\textbf{Scaling data and model size on \textsc{sraven}}. \textbf{A} Compositional generalization measured by OOD accuracy as a function of the number of training problem instances for different widths (scaling embedding and key-query-value dimensions). \textbf{B} Same as A but increasing depth instead of width.}
    \label{fig:sraven-results}
    \vspace{-1em}
\end{figure}
Making the task symbolic instead of presenting images of geometric shapes comes with the advantage of a smaller input space, allowing us to scale to larger model and data sizes in our experiments.
However, it removes the visual representation learning step of discovering the right symbolic abstractions.
We argue that while this is a difficult problem worth studying by itself, there is a major source of difficulty that arises even with access to symbolic abstractions.
As \citet{carpenter_what_1990} argue in their theoretical account of the original Raven test, what makes many tasks difficult is the problem of \textit{finding correspondences} of rules to feature dimensions.

This is best illustrated when considering the example in Figure~\ref{fig:sraven-task}B showing an adapted version of Figure~5 of \citet{carpenter_what_1990}.
When first attempting to solve this task, a test subject might hypothesize that there is a rule guiding the numerosity and orientations per shape.
For example, one might try to find a rule that governs all curved shapes, one rule that governs all lines and one rule that governs all rectangles.
Translated into a symbolic encoding, this is akin to ordering symbols by the shape and trying to find rules along each dimension in this ordering.
However, this approach turns out to not provide a parsimonious answer.
The correct answer can instead be obtained by grouping the objects by their orientations, i.e. by finding a rule that applies to all horizontal objects and a rule that applies to all vertical objects.
In the symbolic encoding this new correspondence of features amounts to a new ordering of the same symbolic encoding where symbols are now ordered by orientation and one needs to find rules guiding the numerosity and shape per orientation.

To model the property of finding correspondences in \textsc{sraven} we similarly allow for permutations of the features.
Specifically, we sample and apply a column-specific permutation of the features when generating each task, such that the entries of the tuples encoding each panel are permuted with a consistent permutation for a given column across all rows.
This makes finding the correct solution more difficult, since a task can no longer be treated as $K$ independent tasks with a single feature and as a result the number of possible hypotheses grows significantly (see Appendix~\ref{appsec:raven-stats} for more details).
Note that in principle, a problem instance might have multiple valid but divergent answers, making the task ill-posed.
In the settings considered here, this happens for less than half a percent of the problem instances.
See Appendix~\ref{appsec:raven-ambiguity} for more details on handling such ambiguities.

\subsection{Results}
\begin{figure}[t]
    \centering
    \includegraphics[width=\textwidth]{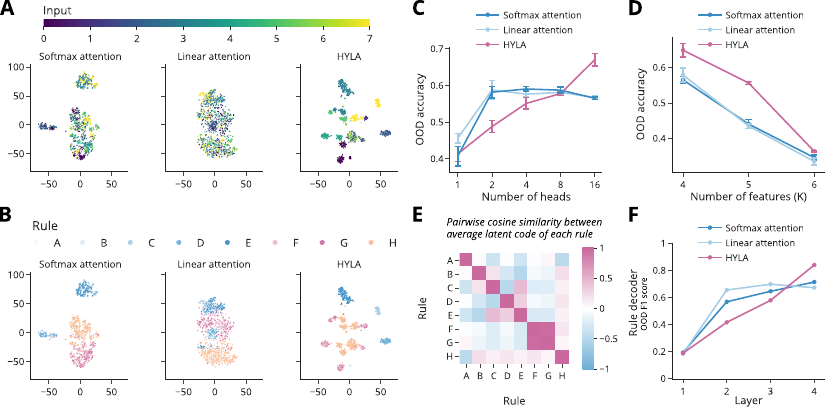}
    \caption{\textbf{Latent code structure of \textsc{sraven}}. \textbf{A} tSNE visualizations of the final layer latent codes for the final response token colored by the magnitude of the predicted target value. \textbf{B} Same as A but colored by the ground-truth rule the model needs to apply to generate the correct prediction. \textbf{C} OOD accuracy for varying numbers of attention heads. For a single head, the hypernetwork mechanism is absent, which hampers OOD generalization. \textbf{D} The difficulty of \textsc{sraven} can be parametrically controlled by varying, $K$, the number of features per panel . \textbf{E} Heatmap showing the pairwise cosine similarity between the average latent code of each rule for HYLA revealing how semantically related rules form clusters. For instance, rule F (addition) and G (difference) are implemented with a very similar latent code, indicating that the same code might be reused by flipping the sign of the operands. \textbf{F} Decoding performance of a logistic regression classifier trained to predict the ground-truth \textsc{sraven} rule based on the latent code at the final response token of training tasks and evaluated on unseen OOD tasks, revealing that the latent code is predictive of the implemented rule.}
    \label{fig:sraven-latent-structure}
    \vspace{-1em}
\end{figure}
\paragraph{Scaling model size and data enables compositional generalization.}
\label{sec:sraven-scaling}
We first evaluate to what extent transformers can compositionally generalize on \textsc{sraven} as we increase the number of problem instances trained on and scale up the models both in terms of width and depth.
Figure~\ref{fig:sraven-results} shows the results of these scaling experiments.
Given sufficient data and model size, all model classes are able to solve 80\% of problem instances created from tasks held-out during training.
The inductive bias of making the value network more expressive by introducing a nonlinearity in HYLA seems to be beneficial for this task.
Especially when trained on fewer problem instances and for small model sizes, it outperforms both linear and softmax attention.

\paragraph{Disrupting the hypernetwork mechanism hurts compositional generalization.}
Varying the number of heads offers another way of manipulating the hypernetwork mechanism of multi-head attention.
Specifically in the case of a single head the mechanism degenerates, only allowing the network to rescale the weights of the sole remaining value projection but no longer allowing it to compose multiple value projections, c.f. equation~\eqref{eq:hya-main}.
In line with this argument, we observe a noticeable decrease in OOD performance in Figure~\ref{fig:sraven-latent-structure}C for the single head case.

\paragraph{The latent code is structured according to the rules of the task.}
We next analyze the latent code for the different models solving the task.
To this end, we collect the attention scores of the response token attending to itself.
Following the perspective of attention as a hypernetwork, we expect latent codes corresponding to the same rule to form clusters.
Indeed, Figure~\ref{fig:sraven-latent-structure}B reveals a strikingly structured code for the final layer across models, with clusters closely matching the rule to be inferred to correctly predict the output for each given task.
In comparison, Figure~\ref{fig:sraven-latent-structure}A colors the points in latent space by the magnitude of the inputs the response token needs to attend to, to make the correct prediction.
This tends to explain clusters more prominently in early layers, as shown in the appendix in Figure~\ref{appfig:sraven-latents} and for a 16 layer softmax transformer in Figure~\ref{appfig:sraven-latents-deep}.

To obtain a better understanding of the semantics of the latent code, we explore the pairwise cosine similarity between average latent codes for each rule across tasks for HYLA.
Figure~\ref{fig:sraven-latent-structure}E shows that semantically related rules like sum/difference can form clusters, suggesting that latent codes might even be reused to solve related task operations.
Finally, we train a logistic regression classifier to predict the ground-truth \textsc{sraven}
rule based on the latent code at the final response token of in-distribution \textsc{sraven} tasks and evaluate it
on unseen OOD tasks.
This reveals that, especially for later layers, the latent code is strongly predictive of the subfunctions performed by the network.
\vspace{-2em}
\begin{wrapfigure}[21]{r}{0.3\textwidth}
    \centering
    \includegraphics[width=0.3\textwidth]{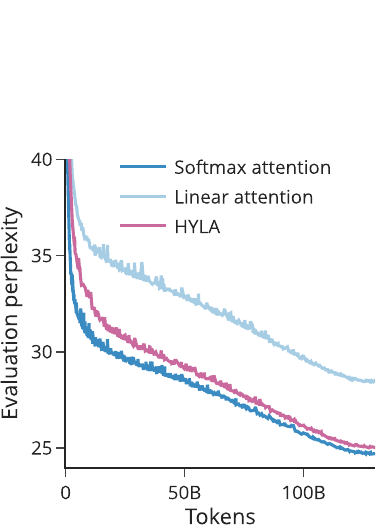}
    \caption{\textbf{Language modeling}. Performance of autoregressively trained decoder-only transformer models with 50M parameters over the course of training on the C4 dataset with 130B tokens.}
    \label{fig:c4}
\end{wrapfigure}
\section{Language modeling}
\label{sec:language}
Multi-head attention has proven especially effective for language modeling.
With language itself being highly compositional, it stands to question whether the implicit hypernetwork mechanism in multi-head attention might play a part in explaining its success.
To shed light on this question, we conducted language modeling experiments and evaluate the performance of HYLA in comparison to linear and softmax attention.
We train decoder-only transformer models with 50M parameters autoregressively for 130 Billion tokens on the C4 dataset \citep{raffel_exploring_2020}.
We find that strengthening the hypernetwork mechanism via HYLA improves performance over linear attention and performs closely to softmax attention despite being a linear attention variant itself.
This is noteworthy considering the importance of softmax in language modeling hypothesized to be due to its role in binding and associative recall problems on which linear attention methods typically struggle \citep{arora_zoology_2023,olsson_in-context_2022, schlag_linear_2021}.
The fact that reinforcing the hypernetwork mechanism implicit in multi-head attention as done by HYLA helps close the gap of linear attention to softmax attention suggests that the hypernetwork mechanism might be of practical relevance for understanding large-scale transformer models.

\section{Related work}
\label{sec:related-work}

\paragraph{The role of multiple heads in attention.}
Prior work that has studied the question of why the attention mechanism benefits from multiple heads has counterintuitively found that it is possible to prune almost all but one head in some layers after training, sacrificing only relatively small amounts of performance \citep{voita_analyzing_2019, michel_are_2019}.
It has therefore been speculated that equipping the attention mechanism with multiple heads primarily aids stability during training \citep{liu_multi-head_2021}.
The hypernetwork perspective of attention offers an alternative account, suggesting that multiple heads create a way of configuring compositional computations in the value network.
Seen from this perspective, the observation of singular heads dominating could point to a connection to the phenomenon of module collapse often observed when training modular systems in practice \citep{shazeer_outrageously_2017, kirsch_modular_2018, rosenbaum_routing_2018, mittal_is_2022}.
\vspace{-1em}
\paragraph{Compositional generalization.}
Consistent with our observation on scaling model size on the \textsc{sraven} task, \citet{hosseini_compositional_2022} find that as pretrained large language models are scaled, their ability to compositionally generalize on in-context semantic parsing tasks improves.
Similarly, outside the in-context learning regime, \citet{furrer_compositional_2021} have demonstrated that pretrained large language models outperform many architectures specialized towards compositional generalization.
Still, to what extent the ability for compositional generalization extends beyond the training distribution remains openly debated~\citep{srivastava_beyond_2023, press_measuring_2023, dziri_faith_2023}.
\vspace{-1em}
\paragraph{Raven-based tasks.}
A number of Raven-inspired tasks have been introduced to assess abstract reasoning capabilities of neural networks \citep{wang_automatic_2015, zhang_raven_2019, barrett_measuring_2018, hu_stratified_2021}.
Common to all these variants, including our version is an underlying generative task model based on a finite number of possible rules that are applied on one or more of the feature dimensions for each panel.
Different from our version, these variants render the problem instances as images using simple geometrical shapes resulting in larger datasets and computational demand.
Still, we argue that the so generated tasks are not necessarily harder than the tasks of \textsc{sraven}, given that \textsc{sraven} models an additional difficulty of \textit{finding correspondences} as detailed in Section~\ref{sec:raven-finding-correspondences} and its difficulty can be controlled parametrically; for instance by increasing the number of features.

\section{Discussion}
\label{sec:discussion}
We have proposed a novel decomposition of multi-head attention, revealing that it can be interpreted as a hypernetwork that composes value networks specific to each key-query pair.
Consistent with this perspective, we find empirically that the attention scores over the heads for each key-query pair form a functionally structured space, identifying reusable subfunctions in two abstract reasoning tasks.
Furthermore, modifying multi-head attention to strengthen the hypernetwork mechanism improves compositional generalization on these tasks.

Adopting the hypernetwork perspective more broadly, multi-head attention can be interpreted as a particular choice for the level of granularity at which the hypernetwork parameterizes the value network: The value networks are \textit{key-query specific }and
are followed by a pooling step that sums the key-query specific outputs over the key index.
In principle, other levels of granularity are possible. 
For instance, the
hypernetwork could parameterize a \textit{query specific} value network which subsumes the aggregation
over the key index.
In the extreme case, it could directly parameterize the full sequence operation potentially facilitating the re-use of sequence-level operations.
This also offers an interesting connection to attention-based graph neural networks~\citep{velickovic_graph_2018, shirzad_exphormer_2023}.
Classically, a non-causal single layer transformer is seen as a fully connected graph where each token corresponds to a node and aggregation is done via attention~\citep{battaglia_relational_2018}.
Our derivation suggests an alternative interpretation where the message function is a hypernetwork that subsumes the attention weights while the aggregation becomes the sum operator.
Given the crucial role that aggregation plays in graph neural networks~\citep{rosenbluth_might_2023,dudzik_asynchronous_2023}, a natural question is whether different pooling operators should be considered more generally for multi-head attention.

\paragraph{Limitations.}
We have focused our analysis on models trained from scratch in order to give us fine-grained control over what data is encountered during training.
Many interesting behaviors emerge in large-scale models pretrained on diverse tasks.
Investigating whether the resulting models form a similarly structured latent code is an interesting avenue for future work.

\paragraph{Ethics statement.}
This paper conducts foundational research aiming to illuminate the mechanisms of the widely used multi-head attention layer.
While we foresee no immediate negative societal impact, we hope that it may improve our understanding of this widely deployed technology.

\paragraph{Reproducibility statement.} To ensure the reproducibility of our work, we are providing the code for all our experiments as part of the supplementary material.
We further expand upon the experimental details explained in the main text in Appendix~\ref{appsec:experimental-details}.

\subsubsection*{Acknowledgements}
We would like to thank Angelika Steger as well as Alexander Meulemans, Johannes von Oswald, Maciej Wołczyk and Owen He for fruitful discussions throughout the development of the project.
This research was supported by an Ambizione grant (PZ00P3\_186027) from the Swiss National Science Foundation and an ETH Research Grant (ETH-23 21-1).

\bibliography{bibliography}
\bibliographystyle{iclr2025}
\appendix
\renewcommand{\thefigure}{A\arabic{figure}}
\setcounter{figure}{0}
\renewcommand{\thetable}{A\arabic{table}}
\setcounter{table}{0}

\begin{figure}
\begin{minipage}{0.48\textwidth}
\begin{algorithm}[H]
\centering
\caption{Multi-head softmax attention}
\label{alg:multihead-attention}
\begin{algorithmic}[1]
\fontsize{8pt}{9pt}\selectfont  %
\Statex \hspace{-\algorithmicindent} \textbf{Input:} $x$
\Statex \hspace{-\algorithmicindent} \textbf{Require:} \texttt{emb\_dim}, \texttt{heads}, \texttt{h\_dim}
    \Statex
    \State $q =$ \textbf{Dense}(shape=(\texttt{heads}, \texttt{h\_dim}), axis=-1)($x$)
    \State $k =$ \textbf{Dense}(shape=(\texttt{heads}, \texttt{h\_dim}), axis=-1)($x$)
    \State $v =$ \textbf{Dense}(shape=(\texttt{heads}, \texttt{h\_dim}), axis=-1)($x$) \\
    \State $a =$ \textbf{einsum}(\texttt{"...qhd,...khd->...hqk"}, $q$, $k$)
    \State $a =$ \textbf{softmax}(a, axis=$-1$) \\
    \State $z =$ \textbf{einsum}(\texttt{"...hqk,...khd->...qhd"}, $a, $v)
    \State
    \State
    \State $y =$ \textbf{Dense}(features=\texttt{emb\_dim}, axis=(-2, -1))($z$)
    \State
    \State \Return $y$
\end{algorithmic}
\end{algorithm}
\end{minipage}
\hfill
\begin{minipage}{0.48\textwidth}
\begin{algorithm}[H]
\centering
\caption{Hypernetwork linear attention}
\label{alg:hyla}
\begin{algorithmic}[1]
\fontsize{8pt}{9pt}\selectfont  %
\Statex \hspace{-\algorithmicindent} \textbf{Input:} $x$
\Statex \hspace{-\algorithmicindent} \textbf{Require:} \texttt{emb\_dim}, \texttt{heads}, \texttt{h\_dim}
    \Statex
    \State $q =$ \textbf{Dense}(shape=(\texttt{heads}, \texttt{h\_dim}), axis=-1)($x$)
    \State $k =$ \textbf{Dense}(shape=(\texttt{heads}, \texttt{h\_dim}), axis=-1)($x$)
    \State $v =$ \textbf{Dense}(shape=(\texttt{heads}, \texttt{h\_dim}), axis=-1)($x$) \\
    \State $a =$ \textbf{einsum}(\texttt{"...qhd,...khd->...hqk"}, $q$, $k$)
    \State \hl{$a =$ \textbf{rms\_norm}(a, axis=$-3$)} \\
    \State $z =$ \textbf{einsum}("...hqk,...khd->...qkd", $a, v$)
    \State \hl{$z =$ \textbf{relu}($z$)}
    \State \hl{$z =$ \textbf{einsum}(\texttt{"...hqk,...qkd->...qhd"}, $a, z$)}
    \State $y =$ \textbf{Dense}(features=\texttt{emb\_dim}, axis=(-2, -1))($z$)
    \State
    \State \Return $y$
\end{algorithmic}
\end{algorithm}
\end{minipage}
\caption{Pseudocode comparing multi-head softmax attention to hypernetwork linear attention. Differences between the two are highlighted in yellow.}
\label{alg:mha-hyla}
\end{figure}

\section{Additional results}

\subsection{Model ablations}
\label{appsec:ablation}
Compared to linear attention, HYLA introduces three modifications: (1) normalization of the attention scores across the head indices using RMSHead, (2) applying the attention-weighted sum operation also to the output projection and (3) inserting a nonlinearity in the value network (c.f. the pseudocode for HYLA shown in Figure~\ref{alg:multihead-attention}).
To delineate the contributions of these modifications, we run an ablation study on the fuzzy logic task and \textsc{sraven} shown in Table~\ref{tab:ablation}.
We find that while simply applying RMSHead to linear attention (Linear Attention + RMSHead) and also the attention-weighted sum operation without nonlinearity (HYLA $-$ nonlinearity) both by themselves can slightly boost performance, the full HYLA model generally performs best.

\paragraph{Increasing depth of the value network.}
To further test our hypothesis that it is the hypernetwork mechanism inside multi-head attention which improves compositional generalization, we explore making the value network configured by the hypernetwork deeper, adding an additional layer as follows,
\begin{align}
\text{HYLA}^+_{q}(\bm{\mathrm{X}})
=& \sum_{k=1}^T \left(\sum_{h=1}^H a_{h,q,k} \; \bm{W}^{\text{out}}_h  \right) \phi \left ( \sum_{h=1}^H a_{h,q,k} \; \bm{W}^{\text{value}'}_h \phi \left ( \sum_{h=1}^H a_{h,q,k} \; \bm{W}^{\text{value}}_h \mathbf{x}_k \right ) \right ),
\end{align}
where $\bm{W}^{\text{value}'}_h \in \mathbb{R}^{D^{\text{key}} \times D^{\text{key}}}$ is a second layer value matrix which increases the number of parameters of this model.
Consistent with our hypothesis, this model also demonstrates strong OOD performance as reported in Table~\ref{tab:ablation}.

\paragraph{Replacing RMSHead normalization with the softmax.}
We perform an additional ablation where we replace the RMSHead normalization of HYLA with the standard softmax that normalizes across keys but not across heads.
As shown in Table~\ref{tab:ablation} this model performs poorly.

\paragraph{Replacing feedforward layers with mixture of experts.}
Sparsely gated mixture of experts (MoE) have been proposed as a replacement for the feedforward layer in the transformer block \citep{shazeer_outrageously_2017}.
In this model, each token is routed to a sparse subset of $k$ out of $E$ experts
$$f_{\mathrm{MoE}}(\bm{x})= \sum_{i=1}^{E}\mathrm{Softmax}(\mathrm{Top\_k}(\bm{W}^{\text{gate}}\bm{x}))\cdot\bm{W}^{\mathrm{up}}_i\phi(\bm{W}^{\mathrm{down}}_i\bm{x}).$$
Intuitively, this could allow individual experts to specialize on specific subtasks that can be composed by the router.
However, it is not well understood if MoEs improve compositional generalization of transformers in practice.
Here, we perform additional experiments where we replace the feedforward layers with MoEs on the \textsc{sraven} and the fuzzy logic task reported in Table~\ref{tab:ablation}.
While we find moderate general improvements, these improvements seem to be similar to other increases of model capacity such as increasing the width/depth as discussed in Section~\ref{sec:sraven-scaling}.

\begin{table}[ht]
\caption{\textbf{Model ablations.} Left column reports the OOD R2 score on the fuzzy logic task studied in Figure~\ref{fig:flogic} for a sequence length of 32 and 70\% of tasks held out from training.
Right column reports the OOD accuracy on the \textsc{sraven} task studied in Figure~\ref{fig:sraven-results} for a 4-layer transformer trained on 20M problem instances. We report the standard error over 3 seeds.}
\label{tab:ablation}
\centering
\begin{tabular}{@{}l@{\hspace{10\tabcolsep}}ll@{}}
\toprule
                              & Fuzzy logic:     & \textsc{sraven}: \\
                              & OOD R2           & OOD accuracy     \\ \midrule
Softmax attention             & $63.28 \pm 2.31$ & $56.56 \pm 1.05$ \\
Linear attention              & $59.89 \pm 5.22$ & $56.30 \pm 1.11$  \\
HYLA                          & $81.13 \pm 7.77$ & $69.13 \pm 1.90$  \\ \midrule
Linear attention $+$ RMSHead    & $68.93 \pm 5.10$  & $59.01 \pm 2.50$  \\
Linear attention $+$ RMSHead $+$ nonlinear value          & $56.91 \pm 4.25$ & $55.38 \pm 0.32$ \\ \midrule
HYLA $-$ RMSHead                & $81.27 \pm 7.88$ & $62.53 \pm 5.73$ \\
HYLA $-$ nonlinearity          & $84.54 \pm 5.32$ & $56.54 \pm 0.21$ \\
HYLA $-$ nonlinearity $-$ RMSHead & $79.42 \pm 7.43$ & $56.33 \pm 1.41$ \\
HYLA $-$ RMSHead $+$ softmax     & $70.98 \pm 7.08$ & $38.21 \pm 2.98$ \\
HYLA $+$ deep value network   & $87.65 \pm 4.54$ & $66.59 \pm 0.08$ \\
\midrule
Softmax attention + MoE & $63.83 \pm 6.07$ & $57.68 \pm 1.31$ \\
Linear attention + MoE & $60.09 \pm 6.16$ & $64.34 \pm 4.3$ \\
HYLA + MoE & $82.37 \pm 6.55$ & $66.94 \pm 3.94$ \\
\bottomrule
\end{tabular}
\end{table}

\begin{figure}[p]
    \centering
    \includegraphics[width=\textwidth]{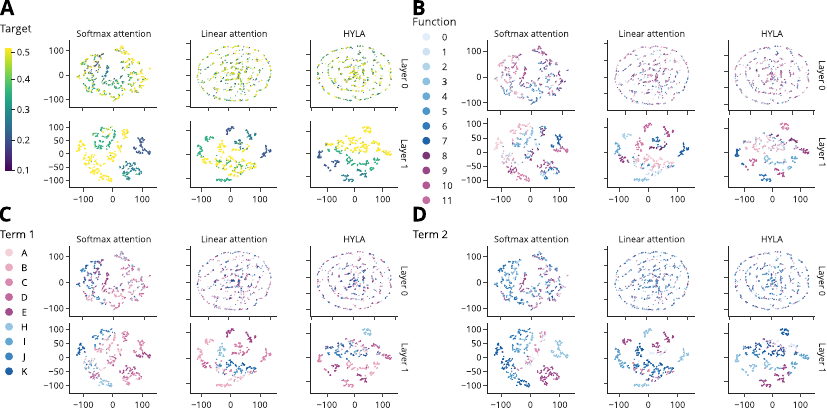}
    \caption{tSNE visualization of the attention scores across the head index for the query token on the fuzzy logic task for a sequence length of 32 and 70\% of tasks held out from training colored according to (A) target label (B) fuzzy logic function, (C) term 1 and (D) term 2.}
    \label{appfig:flogic-latents}
\end{figure}

\begin{figure}
    \centering
    \includegraphics[width=\linewidth]{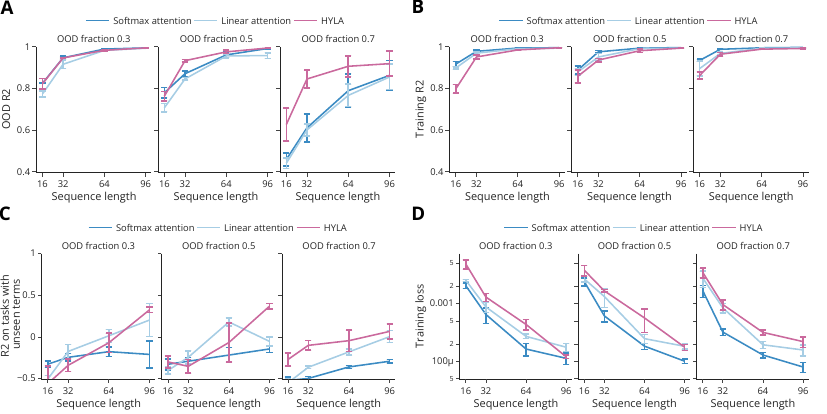}
    \caption{\textbf{Fuzzy logic task performance metrics.} Various task performance metrics on the fuzzy logic task for varying number of in-context examples and fraction of tasks held-out during training. \textbf{A} Coefficient of determination ($R^2$) on tasks consisting of unseen combinations of seen terms (same as Figure~\ref{fig:flogic}\textbf{C})
    \textbf{B} Coefficient of determination ($R^2$) on training tasks.
    \textbf{C} Coefficient of determination ($R^2$) on tasks containing terms not encountered during training. As expected all methods fail in this setting. \textbf{D} Training mean squared error loss.}
    \label{appfig:flogic-seq-len-frac-test}
\end{figure}

\begin{figure}[t]
    \centering
    \includegraphics[width=\textwidth]{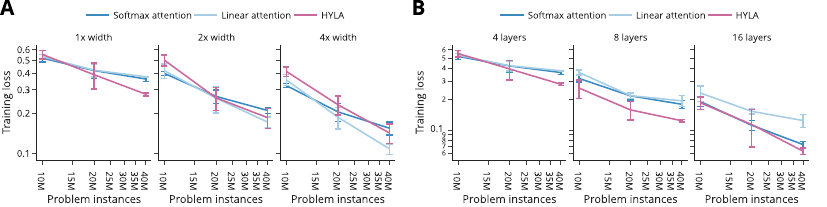}
    \caption{\textbf{Training loss when scaling data and model size on \textsc{sraven}}. \textbf{A} Training loss as a function of the number of training problem instances for different widths (scaling embedding and key-query-value dimensions). \textbf{B} Same as A but increasing depth instead of width.}
    \label{fig:sraven-scaling-loss}
\end{figure}

\begin{figure}[t]
    \centering
    
    \includegraphics[width=0.4\linewidth]{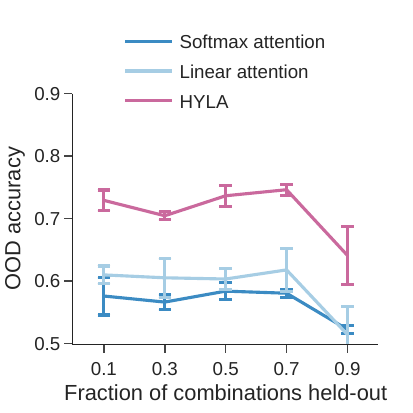}
    \captionsetup{width=0.4\linewidth}
    \caption{\textbf{Varying the fraction of OOD tasks on \textsc{sraven}}. OOD accuracy on \textsc{sraven} as varying fractions of tasks specified by their rule combinations are held-out during training.}
    \label{fig:sraven-frac-ood}
\end{figure}

\begin{figure}[t]
    \centering
    \includegraphics[width=\linewidth]{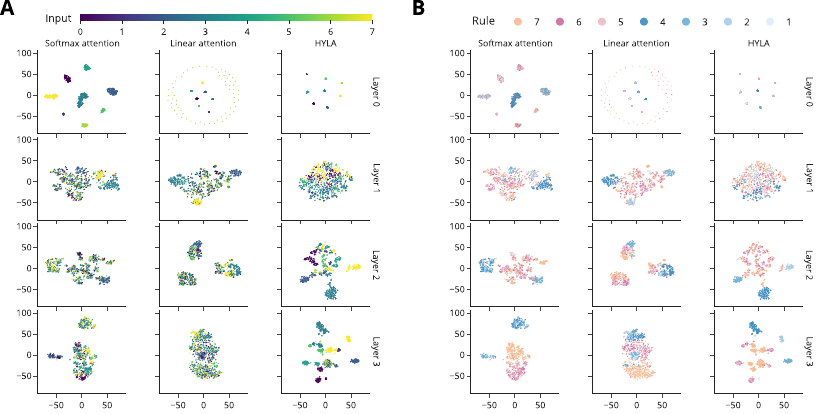}
    \caption{\textbf{Latent code on \textsc{sraven}}. \textbf{A} tSNE visualization of the latent code predicting the final query token colored by the magnitude of the predicted target value for 4-layer transformers trained on 40M problem instances. \textbf{B} Same as A but colored by rule.}
    \label{appfig:sraven-latents}
\end{figure}

\begin{figure}[p]
    \centering
    \includegraphics[height=0.9\textheight]{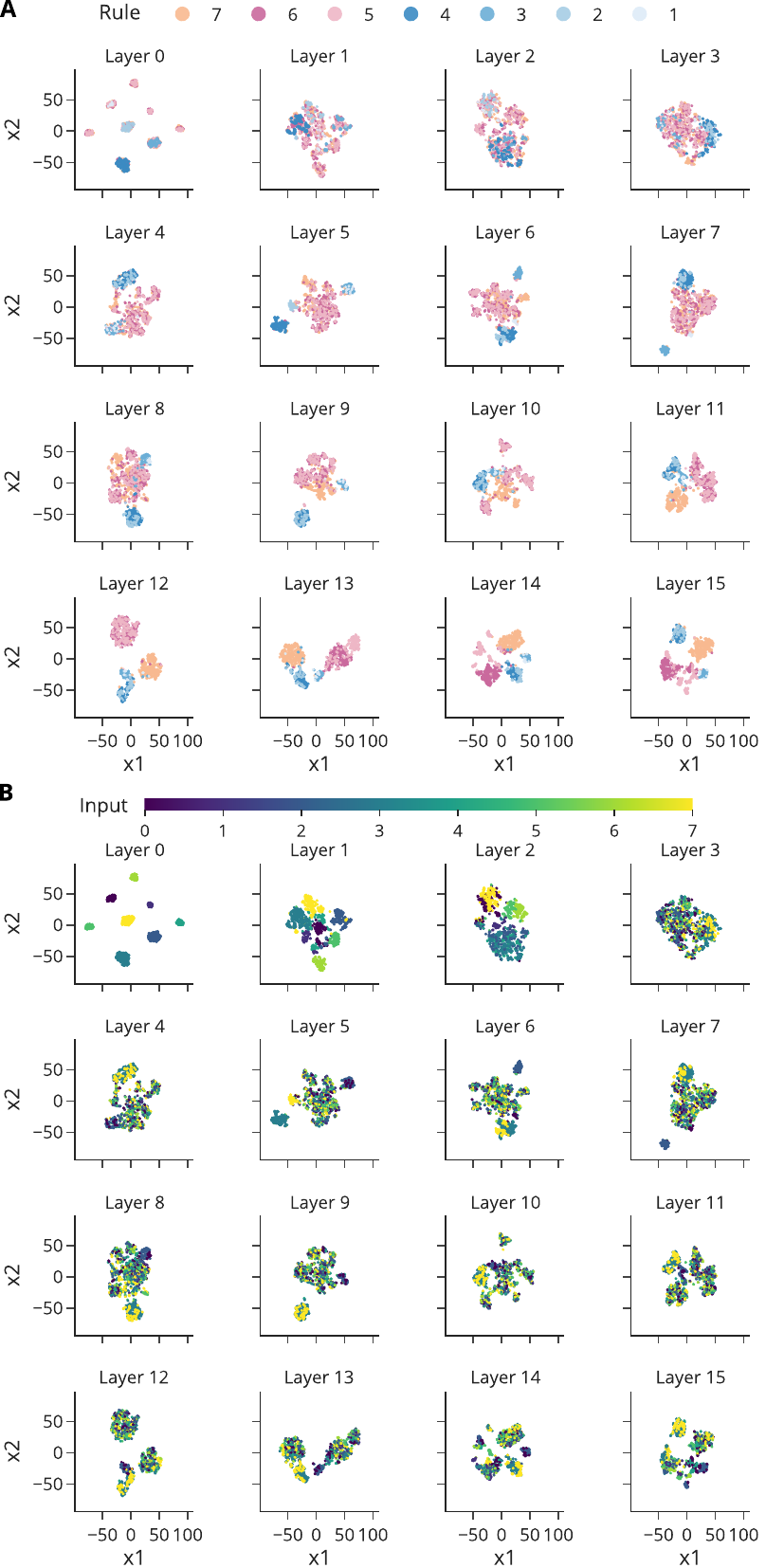}
    \caption{\textbf{Latent code visualization in \textsc{sraven} for a softmax transformer of depth 16}. \textbf{A} tSNE visualization of the latent code predicting the final query token colored by the rule to be inferred for a softmax transformer of depth 16 trained on 40M problem instances. \textbf{B} Same plot as in A but colored by the magnitude of the predicted target value.}
    \label{appfig:sraven-latents-deep}
\end{figure}

\begin{figure}[t]
    \centering
    \includegraphics[width=\linewidth]{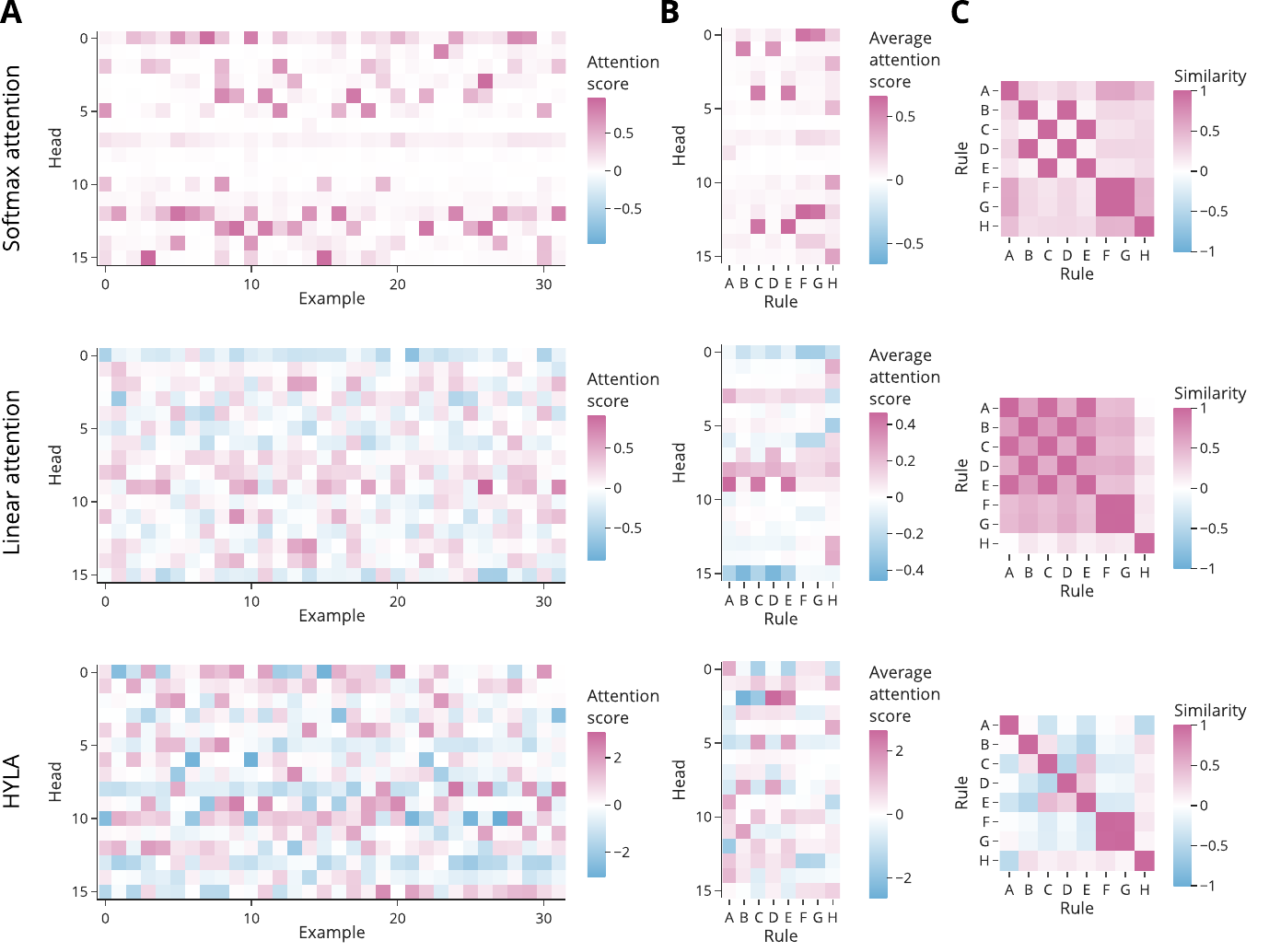}
    \caption{\textbf{Detailed attention scores on \textsc{sraven}}. \textbf{A} Random sample of the attention scores / latent codes for the final key-query pair on OOD \textsc{sraven} tasks in the final layer of a transformer using softmax attention, linear attention or HYLA. \textbf{B} Average attention scores / latent codes per rule. \textbf{C} Pair-wise cosine similarities between all average latent codes per rule. Data from the same models as shown in Figure~\ref{appfig:sraven-latents})}
    \label{fig:sraven-attention}
\end{figure}

\clearpage

\section{\textsc{sraven}}

\subsection{Number of possible tasks and instances}
\label{appsec:raven-stats}

\begin{figure}[p]
    \centering
    \includegraphics[width= \textwidth]{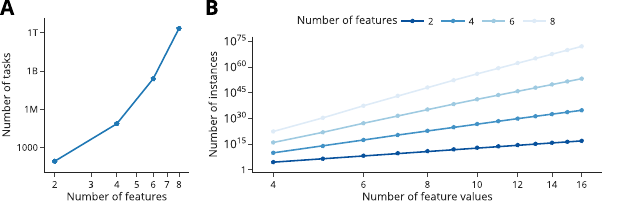}
    \caption{\textbf{A} Number of possible \textsc{sraven} tasks as a function of the number of features ($K$). \textbf{B} Number of possible \textsc{sraven} problem instances as a function of the number of features ($K$) and the number of feature values ($F$).}
    \label{appfig:sraven_stats}
\end{figure}

In the general case, the number of possible tasks in \textsc{sraven} is a product of the number of possible rule combinations $\binom{R + K - 1}{K}$ and the number of possible unique permutations $((K!)^{G-1} - K)$ where $G$ is the number of columns of the grid (in Raven-based tasks typically $G=3$).
Figure~\ref{appfig:sraven_stats}A shows the number of possible \textsc{sraven} tasks and Figure~\ref{appfig:sraven_stats}B shows the number of possible \textsc{sraven} problem instances.
Note specifically that the latter is orders of magnitudes larger than the number of training examples in our experiments for $K=4$, $F=8$.

\subsection{Ambiguity}
\label{appsec:raven-ambiguity}

\begin{table}[p]
\centering
\caption{Monte Carlo estimation of fraction of ambiguous examples over 4096 problem instances $\pm$ standard error. Bold used to denote the setting used for all experiments in this paper.}
\label{apptab:raven-ambiguous}
\begin{tabular}{@{}lll@{}}
\toprule
Number of features ($K$) & Number of feature values ($F$) & Estimated fraction of ambiguous examples \\ \midrule
4                        & 4                              & $0.0642 \pm 0.0038$                      \\
\textbf{4}               & \textbf{8}                     & $\bm{0.0032 \pm 0.0009}$             \\
4                        & 16                             & $0.0005\pm0.0003$                        \\ \bottomrule
\end{tabular}
\end{table}

In principle, a problem instance generated given a set of rules and permutations might have an ambiguous answer.
That is, there is at least one other set of rules and permutations that fit all context panels but prescribe a different final query panel.
One possibility is to check for such examples when generating task instances and reject ambiguous ones.
This process is prohibitively expensive, as it requires a search over the exponentially many possible permutations.
We therefore resort to a Monte-Carlo estimation of the fraction of ambiguous instances.
Table~\ref{apptab:raven-ambiguous} shows the results for various parameterizations of the \textsc{sraven} task.
In the setting we consider here, ($K=4, F=8$), the probabilities are negligible, and we do not take measures to filter out ambiguous examples given the computational overhead it produces in the data generation process.

\subsection{Rules}
\label{appsec:sraven-rules}
\texttt{sraven} contains the following rules:
\begin{enumerate}[leftmargin=*]
    \item \texttt{constant}: Each row consists of a random but fixed integer in the range $1$ to $F$.
    \item \texttt{progression (+1)}: The first element of each row is sampled uniformly at random and incremented by 1 modulo $F$ for each successive column.
    \item \texttt{progression (+2)}: The first element of each row is sampled uniformly at random and incremented by 2 modulo $F$ for each successive column.
    \item \texttt{progression (-1)}: The first element of each row is sampled uniformly at random and decremented by 1 modulo $F$ for each successive column.
    \item \texttt{progression (-2)}: The first element of each row is sampled uniformly at random and decremented by 2 modulo $F$ for each successive column.
    \item \texttt{addition}: Two elements are sampled uniformly at random for each row and added modulo $F$ to obtain the last column.
    \item \texttt{subtraction}: Two elements are sampled uniformly at random for each row and subtracted modulo $F$ to obtain the last column.
    \item \texttt{distribute three}: Three elements are sampled uniformly at random and presented in three independently sampled random permutations for each row.
\end{enumerate}

\clearpage

\section{Experimental details}
\label{appsec:experimental-details}
\paragraph{Architecture.}
For all models we use a standard decoder-only transformer architecture where each block is structured as
\begin{align*}
    \bm{\mathrm{Z}} &= \texttt{MultiHeadAttention}(\texttt{LayerNorm}(\bm{\mathrm{X}})) + \bm{\mathrm{X}} \\
    \bm{\mathrm{Y}} &= \texttt{FeedForward}(\texttt{LayerNorm}(\bm{\mathrm{Z}})) + \bm{\mathrm{Z}}.
\end{align*}
\texttt{MultiHeadAttention} is either linear attention, softmax attention or hypernetwork linear attention (HYLA) and uses T5-style relative positional embeddings \citep{raffel_exploring_2020}.
For the \texttt{FeedForward} layer, we employ a standard single-hidden layer MLP with GeLU nonlinearity applied to each position in the sequence independently.
For our language modeling experiments we use rotary positional encodings \citep{su_roformer_2024}.

\paragraph{Tokenization.}
For the fuzzy logic task we directly use the concatenated input examples as tokens.
Similarly for \textsc{sraven} we use one-hot encodings of the integer inputs as the input tokens.
In both cases the input tokens are mapped into the embedding space of the transformer using a fully connected dense layer and another fully connect dense layer is used to produce the output logits.
For the language modeling experiment we use the default sentencepiece tokenizer \citep{kudo_sentencepiece_2018} of NanoDo \citep{liu_nanodo_2024} with a vocabulary size of $32000$ and use the transposed embedding matrix to produce the logits.

\paragraph{Optimization.}
We use the AdamW optimizer \citep{loshchilov_decoupled_2019} with linear warmup starting from a learning rate of $0$ followed by a cosine decay learning rate scheduler \citep{loshchilov_sgdr_2017} that decays the learning rate to $0.1$ times the base learning rate.
We exempt biases and LayerNorm parameters from weight decay.
In the fuzzy logic task we use the mean-squared error on the logits of the query token whereas in \textsc{sraven} we use the softmax cross-entropy loss on the logits of all query tokens.
The language modeling task uses an autoregressive softmax cross-entropy loss with causal masking applied to the attention mechanism as well as mixed-precision training.

\paragraph{Hyperparameters.}
For all tasks and models we perform a grid search over the learning rate, weight decay and warmup steps.
We report the search grid as well as all other hyperparameters in Table~\ref{apptab:hyperparameters}.

\begin{table}[htb]
\centering
\caption{Hyperparameters for all tasks and methods. Lists of values indicates that a grid search over all points contained in the list have been conducted and for each method the optimal combination has been picked. The width factor multiplies the embedding dimension, key-query-value dimension and MLP dimension.}
\label{apptab:hyperparameters}
\begin{tabular}{@{}llll@{}}
\toprule
Parameter   & Fuzzy logic  & \textsc{sraven} & Language modeling     \\ \midrule
\texttt{batch\_size}   & $128$    & $128$ & $256$      \\
\texttt{num\_layers}   & $2$    & $4$ / $8$ / $16$ & $6$      \\
\texttt{width\_factor} & $1$    & $1$ / $2$ / $4$ & $1$      \\
\texttt{emb\_dim}   & $128$    & $128$    & $512$      \\
\texttt{kqv\_dim}   & $16$    & $64$    & $64$      \\
\texttt{mlp\_dim}   & $256$    & $256$    & $2048$      \\
\texttt{num\_heads}   & $8$    & $16$    & $8$      \\
\texttt{learning\_rate}  & $[0.001, 0.003]$ & $[0.0003, 0.001]$ & $[0.001, 0.003]$ \\
\texttt{weight\_decay}  & $[0.03, 0.1]$ & $[0.1, 0.3]$  & $[0.1, 0.03]$      \\
\texttt{warmup\_steps}  & $100$    & $[1000, 3000]$ & $1000$      \\ \bottomrule
\end{tabular}
\end{table}

\clearpage

\section{Additional details}

\subsection{Compute resources}
\label{appsec:compute-resources}
We used a Linux workstation with two Nvidia RTX 3090 GPUs with 24GB of memory each for development and conducted hyperparameter searches and experiments using 1 Linux server with 4 Nvidia RTX 3090 GPUs as well as a Slurm cluster equipped with Nvidia RTX 4090 GPUs.
With a single GPU of these two GPU models, a single run of the fuzzy logic task takes around $2-3$ minutes, a single run on \textsc{sraven} between $20-200$ minutes.
For the language modeling experiments, we used 16 Cloud TPU v5e with a complete run taking 72-100 hours.
In total, it takes around 24 GPU hours to reproduce all fuzzy logic results on our hardware, around 10 GPU days to reproduce all our \textsc{sraven} results and approximately 163 TPU days to reproduce our results on C4.

\subsection{Software and libraries}
\label{sec:software}
For the results obtained in this paper we built on free and open-source software.
We implemented our experiments in Python using JAX \citep[][Apache License 2.0]{bradbury_jax_2018}, Flax \citep[][Apache License 2.0]{heek_flax_2023}, NanoDo \citep[][Apache License 2.0]{liu_nanodo_2024} and the Deepmind Jax Ecosystem \citep[][Apache License 2.0]{babuschkin_deepmind_2020}.
We utilized WandB \citep[][MIT license]{biewald_experiment_2020} to monitor the progress and results of experiments, and Plotly  \citep[][MIT license]{inc_collaborative_2015} for generating the plots.

\end{document}